\documentclass{llncs}
\usepackage{makeidx}  
\usepackage{amssymb}
\usepackage{graphicx}
\usepackage{epsfig}
\usepackage{amsmath}
\usepackage{array}
\usepackage[T1]{fontenc}
\newcolumntype{M}[1]{>{\centering\arraybackslash}m{#1}}
\begin{document}
\mainmatter              
\title{Deep Reinforcement Learning for Surgical Gesture Segmentation and Classification}
\titlerunning{RL for Surgical Gesture Segmentation and Classification}
%
%
\author{Daochang Liu \and Tingting Jiang}
\authorrunning{Daochang Liu and Tingting Jiang} 
\tocauthor{Daochang Liu and Tingting Jiang}
\institute{National Engineering Lab for Video Technology, Cooperative Medianet\\
Innovation Center, School of EECS, Peking University, Beijing 100871, China\\
\email{\{daochang, ttjiang\}@pku.edu.cn}}
\maketitle              

\begin{abstract}
Recognition of surgical gesture is crucial for surgical skill assessment and efficient surgery training. Prior works on this task are based on either variant graphical models such as HMMs and CRFs, or deep learning models such as Recurrent Neural Networks and Temporal Convolutional Networks. Most of the current approaches usually suffer from over-segmentation and therefore low segment-level edit scores. In contrast, we present an essentially different methodology by modeling the task as a sequential decision-making process. An intelligent agent is trained using reinforcement learning with hierarchical features from a deep model. Temporal consistency is integrated into our action design and reward mechanism to reduce over-segmentation errors. Experiments on JIGSAWS dataset demonstrate that the proposed method performs better than state-of-the-art methods in terms of the edit score and on par in frame-wise accuracy. Our code will be released later.
\end{abstract}
\section{Introduction}
Joint surgical gesture segmentation and classification is fundamental for objective surgical skill assessment and for improving efficiency and quality of surgery training \cite{ahmidi2017dataset}. The goal is to segment robotic kinematic data or video sequence and to classify segmented pieces into surgical gestures, such as {\itshape reaching for the needle}, {\itshape orienting needle} and {\itshape pushing needle through the tissue}, etc.

Variant temporal models have been exploited in prior works on surgical gesture segmentation and classification. One branch of works has been based on hidden Markov models (HMMs) \cite{tao2012sparse,sefati2015learning,varadarajan2009data}, differing from each other in how the emission probability is modeled. HMM-based methods assume that gesture label at frame $t$ is only conditioned on previous frame $t-1$, leaving long-term dependency unconsidered. Another branch has been based on conditional random fields (CRFs) \cite{tao2013surgical,lea2015improved,lea2016learning} and their extensions, which obtains the gesture sequence by minimizing an overall energy function. Although these methods capture temporal patterns by the pairwise potentials in their energy functions, they produce severe over-segmentation and therefore suboptimal segmental edit scores. In recent years, a third branch using deep learning has set new benchmarks for this task. Recurrent neural networks, in particular LSTMs, were applied in \cite{dipietro2016recognizing}. A memory cell is maintained in LSTM to remember and forget action changes over time. \cite{lea2016segmental} proposed a spatiotemporal CNN, in which the spatial component described relationships of objects in the scene, and a long temporal convolutional filter captured how the relationships change temporally. Thereupon, \cite{Lea_2017_CVPR,lea2016temporal} went further and built a hierarchical encoder-decoder network called Temporal Convolutional Network (TCN) composed of long temporal convolutional filters, upsampling/downsampling layers, and normalization layers. In spite of the promising performance improvement achieved, these methods are only driven by frame-wise accuracy due to their cross-entropy training loss.

Unlike prior works, we propose an essentially different deep reinforcement learning approach for joint surgical gesture segmentation and classification, which is driven by both frame-wise accuracy and segment-level edit score. Reinforcement learning has gained remarkable success recently in domains like playing Go \cite{silver2016mastering}, Atari games \cite{mnih2015human}, and anatomical landmark detection in medical images \cite{ghesu2016artificial}, etc. However, reinforcement learning has not been applied in surgery gesture segmentation in existing works. We formulate the task as a sequential decision-making process and train an agent to operate in a human-like manner. The agent looks through the surgical data sequence from the beginning, segment the sequence step by step and classify frames simultaneously. To highlight, our agent learns a strategical policy---skim fast in the middle of segments and examine attentively at segment boundaries, which resembles human intelligence. Additionally, current deep learning methods like RNN and TCN handle temporal consistency {\itshape implicitly} by memory cells or temporal convolutions. On the contrary, we enforce temporal consistency {\itshape explicitly} by the design of action and reward. The reward consists of two terms that guide the agent to high accuracy and high edit score respectively. To combine reinforcement learning with the hierarchical representation learned by deep neural networks, features extracted by TCN are utilized as powerful state representation for the agent. The proposed method is tested on the suturing task of the JIGSAWS dataset \cite{ahmidi2017dataset,gao2014jhu}. Experiments show that our method outperforms state-of-the-art methods in terms of edit score. In summary, our contributions are three-fold: 
\begin{enumerate} 
  \item[--] Joint surgical gesture segmentation and classification is formulated, to our best knowledge for the first time, as a sequential decision-making problem using deep reinforcement learning. 
  \item[--] The two evaluation metrics of frame-level accuracy and segment-level edit score are both incorporated into the rewarding mechanism.
  \item[--] Our method outperforms state-of-the-art methods in terms of edit score while retaining comparable frame-wise accuracy.
\end{enumerate}

\section{Preliminary}

Reinforcement learning (RL) is a computational approach to decision-making problems with definite goals \cite{sutton1998reinforcement}, where an artificial agent learns its policy from interactions with the environment. At each time step, the agent observes the state of environment and selects an action accordingly, which in turn affects the environment. The agent earns a numerical reward for each action and updates its policy to maximize future reward. This sequential decision-making process is formalized as a Markov Decision Process (MDP) \cite{sutton1998reinforcement} $\mathcal{M}:=(\mathcal{S},\mathcal{A},\mathcal{P},\mathcal{R},\gamma)$, where $\mathcal{S}$ is a finite set of states, $\mathcal{A}$ is a finite set of actions, $\mathcal{P}:\mathcal{S}\times\mathcal{A}\times\mathcal{S}\rightarrow[0,1]$ denotes state transition probabilities, $\mathcal{R}:\mathcal{S}\times\mathcal{A}\rightarrow \mathbb{R}$ denotes a reward function for each action performed in certain state, and $\gamma \in [0,1]$ is the discount factor balancing between immediate and long-term reward. The agent learns from experience to optimize its policy $\pi:\mathcal{S}\times\mathcal{A}\rightarrow[0,1]$, which is stochastic in general. The goal is to maximize discounted future reward accumulated from current time step to the end of learning episode.

\section{Proposed Method}

The input to our model is a data sequence $\{x_t\}$, which can be either visual features extracted from surgery video frames or kinematic data frames collected from surgical robots, together with its ground truth gesture label sequence $\{y_t\}$. Each $x_t \in \mathbb{R}^{n_x}$ and each $y_t \in \{1,2,\dots,n_y\}$, where ${n_x}$ is the dimensions of features, ${n_y}$ is the number of gesture classes, and $1 \leq t \leq n_t$. Note that the number of frames $n_t$ may differ between each data sequence.

\begin{figure}[t]
\centering
\epsfig{file=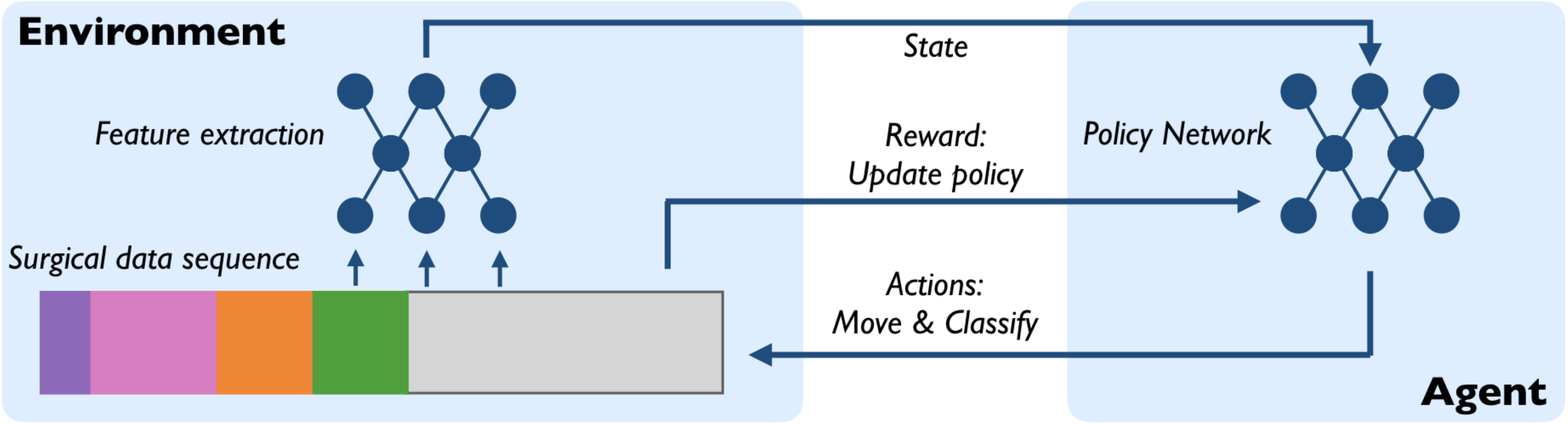,width=11cm,height=2.5cm}
\caption{Overview diagram of the proposed method. An agent perceives the environment, selects an action, and update its policy to maximize future rewards.}
\label{fig:f1}
\end{figure}

We propose to model joint surgical gesture segmentation and classification as a sequential decision-making problem, illustrated in Fig. \ref{fig:f1}. An agent is built to interact with the visual or kinematic data sequence, i.e., the environment. Initially positioned at the beginning of data sequence, the agent selects a proper step size and moves forward at each time step, concurrently classifying frames stepped over. Following the standard paradigm of reinforcement learning, we formalize the task as a Markov Decision Process. The action, the state and the reward of our proposed MDP model are detailed as follows. 

\textbf{Action.} The action of the proposed model includes two subactions, which are to decide how far to move forward and to choose which class label to give. The action set is defined as:
\begin{equation}
  \mathcal{A} := \mathcal{K} \times \mathcal{C}
\end{equation}
where $\mathcal{K}$ represents a predefined set of optional step sizes, and $\mathcal{C}$ represents the set of gesture classes. From the start, the agent keeps selecting an action pair $(k,c)$ from action set $\mathcal{A}$ to walk through and classify the data sequence until reaching the end. Step size set $\mathcal{K}$ is defined to contain one small step and one large step: $\mathcal{K}:=\{k_s,k_l\}$. This binary design enables the agent to alter its step size based on the confidence in the gesture label to give. The agent can adopt the smaller step when the state is not discriminative enough such as at the boundaries between gestures, and adopt the larger step otherwise. At each action, the $k$ frames stepped over by the agent are labeled with the same class $c$, explicitly enforcing temporal consistency.

\textbf{State.} Raw features $\{x_t\}$ are challenging for the agent to fully understand the surgical activity. Therefore we define the state to be a combination of high-level representation and other auxiliary information, which assists the agent to make a better decision. We utilize TCN \cite{Lea_2017_CVPR,lea2016temporal} to extract such high-level representation from raw features.

The state observed at frame $t$ is the concatenation of current and future feature vectors extracted by TCN, gesture transition probabilities from a language and duration model \cite{richard2016temporal}, and a one-hot vector, which is formalized below:
\begin{equation}
  s^t := (s_{tcn}^t, s_{tcn}^{t+k_s}, s_{tcn}^{t+k_l}, s_{trans}, s_{hot})
\end{equation}
where $s_{tcn}^t, s_{tcn}^{t+k_s}, s_{tcn}^{t+k_l}$ are respectively TCN features at the current frame, $k_s$ frames later and $k_l$ frames later, $s_{trans}$ are probabilities of transition into each gesture computed from a statistical language model, $s_{hot} \in \{0,1\}^{n_y}$ is a one-hot vector such that the gesture class given by last action is 1 and all others are 0.

{\itshape Statistical Language and Duration Model.} Similar to \cite{richard2016temporal}, we use a statistical model to describe the length and contextual pattern of gestures. Specifically, we use Gaussian distributions for gesture durations and a bigram language model for gesture transitions, assuming the gesture class depends only on one previous class. The statistical model is formalized as:
\begin{equation}
p(i\,|\,j,l) := \begin{cases}
\frac{N(j,i)}{N(j)} \cdot CDF_j(l) &\text{if $i \neq j$}\\
1-CDF_j(l) &\text{if $i=j$}
\end{cases}
\end{equation}
where $p(i\,|\,j,l)$ is the probability of transition from gesture $j$ to gesture $i$ given $l$---how many frames the agent has stayed in gesture $j$, $N(j,i)$ and $N(j)$ are respectively occurrence counts of ordered gesture pair $(j,i)$ and gesture $j$ alone in training data, $CDF_j$ stands for the cumulative distribution function of a Gaussian distribution modeling the length of gesture $j$. The Gaussian distributions are parameterized by maximum likelihood estimation using training data. Then $s_{trans}$ is set as the probabilities of transition to each gesture, given the gesture class of last action and how long the agent has stayed in this gesture.

\textbf{Reward.} The reward is numerical feedback for each action performed by the agent. Given action pair $(k,c)$ performed at frame $t$, the reward in our MDP model is designed as: 
\begin{equation}
  r(s^t,(k,c)) := \alpha k - \sum_{t'=t}^{t+k-1} {\bbbone(y_{t'} \neq c)}
\end{equation}
where the first term encourages the agent to adopt the larger step, the second term penalizes the errors caused by this action, and $\alpha$ is a weight parameter balancing the two terms. The two evaluation metrics of accuracy and edit score are both incorporated in this reward inherently. While the second term serves as straightforward guidance for the agent to achieve high frame-wise accuracy, the first term is crucial for a good edit score at segment-level. Preference for the larger step can mitigate the jittering between gestures and therefore can reduce over-segmentation errors, which is validated in our experiments section. 

\textbf{Policy Learning.} We use a multilayer perceptron (MLP) to model the policy, whose input layer has the same number of units as the dimensions of the state, and output layer has the same number of units as the dimensions of action space. The policy network takes environment state as input and output a distribution over action space. With the MDP process well defined, any standard reinforcement learning method can be applied to policy learning. We choose Trust Region Policy Optimization (TRPO) \cite{schulman2015trust} since it is theoretically guaranteed to improve the policy monotonically. 
%
%

\section{Experiments}

We evaluate the proposed method on JIGSAWS \cite{ahmidi2017dataset,gao2014jhu}, a public benchmark dataset recorded using the {\itshape da Vinci} surgical system. We use the video and kinematic data from the suturing task, which contains 39 sequences performed by eight subjects with varying skill levels. For video data, we use features extracted from each frame image by a spatial CNN \cite{lea2016segmental}, which is consistent with TCN \cite{lea2016temporal}. As for kinematic data, we pick the same subset of features as in \cite{lea2016temporal}, which are position, velocity and gripper angle for both slave manipulators.

\textbf{Experiments Setup.} The standardized leave-one-user-out (LOUO) evaluation setup of JIGSAWS is followed in our experiments. Trials performed by a single user are left out as testing set while all remaining trials are used as training set, resulting in 8-fold cross-validation. Since our RL based method is inherently stochastic, we train the TCN for state feature extraction five times, train the agent using TRPO three times, and test on each data sequence ten times, with $5*3*10=150$ runs in total. Results are averaged over these 150 runs. We also include an ablation study to measure the impact of each component of the state.

We include three evaluation metrics: accuracy, edit score, and F1 score. Accuracy is the percentage of correctly labeled frames, measuring the performance at the frame level. Edit score is the normalized Levenshtein distance between predicted gesture sequence and ground truth, measuring the performance at the segment level, which is between 0 and 100 (the higher the better). F1 score is introduced for this task in \cite{Lea_2017_CVPR}. Each predicted gesture segment is considered to be true or false positive according to whether its Intersection over Union (IoU) with respect to the corresponding ground truth segment is above a threshold. Then F1 score is the harmonic mean of the precision and recall: $F1=\frac{2*precision*recall}{precision+recall}$.

\textbf{Implementation Details.} The proposed model is implemented with Python and {\itshape OpenAI Baselines} library \cite{baselines}. The policy network is of one hidden layer and 64 hidden units. We set $k_s$ to be the minimum gesture length in training set, and $k_l$ to be the minimum of mean gesture lengths for each class, which are 4 and 21 frames for example. The discount factor $\gamma$ and the reward weight $\alpha$ are set to 0.9 and 0.1 respectively in experiments. Besides, we re-implement the TCN for state feature extraction using PyTorch with several minor changes, which can be regarded as a baseline for our RL method. The convolutional layers in the decoder of TCN are replaced with transposed convolutional layers. And due to the data is highly imbalanced, we use weighted cross-entropy as the training loss of TCN instead. Activations before the last fully-connected layer are used as state features, which is of 32 dimensions. Our code will be released later.

\textbf{Result.} Experiment results on both video and kinematic data are shown in Table. \ref{table:t1}. Our RL based approach is compared to the original TCN, the modified TCN and several other recent works. All results of prior works are excerpted from \cite{lea2016temporal}. Compared to existing works, the proposed method achieves higher edit score and F1 score at a negligible cost of accuracy. We present the result of ablation study on the state design as well. Each of following components is removed from the state to justify its necessity: 1) all TCN features 2) TCN features at future frames 3) transition probabilities from the statistical model. Results show that all these components are required to achieve the best performance. And the high-level representation extracted by TCN is the most important one.

\begin{table}[t]
\centering
\caption{Results on the suturing task of JIGSAWS. F1@\{10,25,50\} stands for F1 score with the IoU threshold set to 10\%, 25\% and 50\%. TCN* stands for our re-implemented version of TCN with several minor changes for state feature extraction. The four entries at the bottom are results of our RL method with partial state or full state.}
\label{table:t1}
\scalebox{0.9}{
\begin{tabular}{M{2.4cm}|M{1cm} M{1cm} M{2.4cm}|M{1cm} M{1cm} M{2.4cm}}
\hline
 \multicolumn{1}{c|}{ } & \multicolumn{3}{c|}{Video} & \multicolumn{3}{c}{Kinematic} \\
 Method & Acc & Edit & F1@\{10,25,50\} & Acc & Edit & F1@\{10,25,50\} \\ 
\hline
 SD-SDL \cite{sefati2015learning} & - & - & - & 78.6 & 83.3 & - \\ 
 Bidir LSTM \cite{dipietro2016recognizing} & - & - & - & 83.3 & 81.1 & - \\ 
 LC-SC-CRF \cite{lea2016learning} & - & - & - & \textbf{83.4} & 76.8 & - \\ 
 Seg-ST-CNN \cite{lea2016segmental} & 74.7 & 66.6 & - & - & - & - \\
 TCN \cite{lea2016temporal} & 81.4 & 83.1 & - & 79.6 & 85.8 & - \\  
 TCN* & \textbf{81.71} & 86.63 & 91.0, 89.5, 82.0 & 82.57 & 86.58 & 90.5, 89.3, \textbf{82.4} \\
\hline  
 RL (no tcn) & 32.35 & 15.04 & 13.1, 6.1, 3.4 & 32.90 & 17.21 & 14.8, 7.5, 4.6 \\
 RL (no future) & 80.84 & 81.58 & 87.5, 85.6, 77.4 & 81.48 & 81.07 & 86.5, 84.6, 77.4 \\
 RL (no trans) & 81.32 & 87.30 & 91.5, 90.1, 81.8 & 82.06 & 87.12 & 90.7, 89.1, 81.8 \\
 RL (full) & 81.43 & \textbf{87.96} & \textbf{92.0}, \textbf{90.5}, \textbf{82.2} & 82.07 & \textbf{87.86} & \textbf{91.1}, \textbf{89.5}, 82.3 \\  
\hline  
\end{tabular}}
\end{table}

\begin{figure}[t]
\centering
\epsfig{file=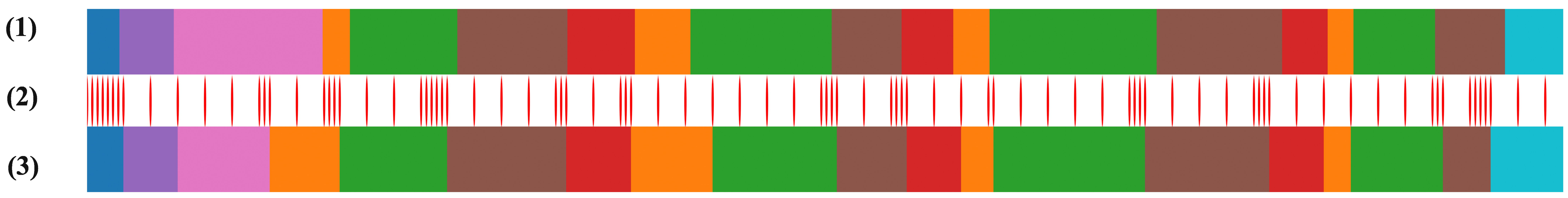,width=\textwidth}
\caption{Prediction example: (1) ground truth sequence (2) step history of the agent (3) predicted sequence. Each color corresponds to a gesture. Our agent learns to skim fast in the middle of gestures and examine cautiously at the boundaries.}
\label{fig:f2}
\end{figure}

A prediction example produced by our agent is provided in Fig. \ref{fig:f2}. We plot the history of steps of the agent, finding it interesting that the agent tends to select the larger step in the middle of gestures and the smaller step at the boundaries. Such behavior pattern verifies our intuitions on the action and reward design.

\textbf{Does the Larger Step Really Benefit?} To further validate our reward design that encourages the larger step, we complete a comparative experiment on the step size. We set the step size set $\mathcal{K}$ to contain only a single option, and set this option to 1, 2, 4, 8, 16, 32 frames. From the result in Table. \ref{table:t2}, the larger step the agent can take, the higher edit score is achieved unless the step is excessively large. But larger steps such as 8, 16 and 32 degrade the accuracy considerably. Our binary design of step sizes achieves the promising result on both metrics.

\begin{table}[t]
\centering
\caption{Experiments on step size}
\label{table:t2}
\scalebox{0.75}{
\begin{tabular}{M{2.4cm}|M{1.5cm} M{1.5cm}|M{1.5cm} M{1.5cm}}
\hline
 \multicolumn{1}{c|}{ } & \multicolumn{2}{c|}{Video} & \multicolumn{2}{c}{Kinematic} \\
 Step Size & Acc & Edit & Acc & Edit \\ 
\hline
 1 & 81.65 & 80.09 & 82.31 & 80.12 \\  
 2 & 81.67 & 83.46 & 82.24 & 82.47 \\  
 4 & 81.53 & 86.04 & 82.16 & 85.35 \\  
 8 & 80.95 & 87.56 & 81.91 & 87.27 \\  
 16 & 79.12 & 88.28 & 79.81 & 88.13 \\  
 32 & 72.94 & 84.05 & 72.86 & 84.04 \\  
\hline  
 4 \& 21 & 81.43 & 87.96 & 82.07 & 87.86 \\  
\hline  
\end{tabular}}
\end{table}

\section{Conclusion and Future Work}

In this work, we proposed a novel method based on deep reinforcement learning for joint surgical gesture segmentation and classification. An artificial agent is trained to act in a human-like manner. By the state, the action and the reward formulation, temporal consistency is explicitly stressed and over-segmentation errors are reduced. The proposed method outperforms the state-of-the-art on JIGSAWS dataset in terms of edit score, while retaining comparable frame-wise accuracy. Future work can be made in following two aspects: 1) developing the step size options into a continuous set 2) combining the state feature extraction network and the policy network for an end-to-end model.

\textbf{Acknowledgement.} This work was partially supported by National Basic Research Program of China (973 Program) under contract 2015CB351803 and the Natural Science Foundation of China under contracts 61572042, 61390514, 61527804. We also acknowledge the high-performance computing platform of Peking University for providing computational resources. Thanks to Dingquan Li for his valuable comments and inspiration.
%
%
\bibliographystyle{splncs}
\bibliography{ref}

\end{document}